\documentclass[11pt, letterpaper]{article}

\usepackage{amsmath, amssymb, amsthm}

\newtheorem{lemma}{Lemma}
\newtheorem{proposition}{Proposition}
\newtheorem{assumption}{Assumption}

\usepackage[margin=1in]{geometry}
\usepackage{graphicx}
\usepackage{booktabs}
\usepackage{float}
\usepackage{multirow}
\usepackage{multicol}
\usepackage{subcaption}
\usepackage{placeins} 
\usepackage{hyperref}
\usepackage{xcolor}
\usepackage{algorithm}
\usepackage{algpseudocode}
\usepackage{authblk}

\usepackage[accsupp]{axessibility}

\title{SCOPE: Scalable Training-Free Semantic Coreset Selection for Long-Tailed Federated Learning}

\author[1]{Md Anwar Hossen\textsuperscript{*}}
\author[2]{Nathan R. Tallent}
\author[2]{Luanzheng Guo}
\author[1]{Ali Jannesari}

\affil[1]{Iowa State University, Ames, Iowa, USA}
\affil[2]{Pacific Northwest National Laboratory, Richland, WA, USA}

\affil[ ]{\texttt{$^1$\{manwar, jannesar\}@iastate.edu}, \texttt{$^2$\{Nathan.Tallent, lenny.guo\}@pnnl.gov}}
\affil[ ]{\vspace{0.5em} \textsuperscript{*}Corresponding author}

\date{} 
\begin{document}

\maketitle
\begin{abstract}
Scientific discovery increasingly relies on federated learning over data from high-resolution instruments, and these data are strongly non-IID and long-tailed. Coreset selection could shorten training, but existing federated coreset methods either prune with local heuristics that ignore the global class distribution, or first train a warm-up model at every client and upload class centroids of its features. This makes coreset selection a distributed bottleneck that consumes GPU time at every client, grows with the model and the number of clients, and must be repeated for every architecture. We introduce SCOPE, named for Semantic Coreset using Orthogonal Projection Embeddings, a scalable, training-free federated coreset framework for long-tailed data. Each client scores its samples once with a frozen, compact vision-language model and uploads scalar statistics per class. The server merges these into a global class profile, which lets each client remove anomalies and prune redundant samples only from classes that are locally abundant but globally common. SCOPE therefore requires no warm-up training, needs a single exchange with the server, and uploads a payload that depends only on the number of classes. On GPU nodes of an HPC cluster, with four datasets, including scientific micrographs, and six baselines, SCOPE selects its coreset up to $7.7\times$ faster and uploads up to $293\times$ less data than the feature-based baseline. As the number of clients grows from 10 to 500, SCOPE's time before training grows by fewer seconds than any baseline's, and at 500 clients the baselines need 1.7 to $3.6\times$ SCOPE's time. Trained on the coreset, the global model reaches 98\% of the full-data accuracy in 18 to 42\% fewer global rounds than full-data training, with 8 to 21\% less client computation over 200 rounds.

\vspace{1em}
\end{abstract}

\section{Introduction}
\label{sec:intro}
Scientific discovery increasingly relies on training AI/ML models over ad hoc federations of highly skewed datasets. These federations arise from collaborative projects built around high-resolution instruments, each with its own installation and purpose, that continuously produce experimental data ranging from microscopy images to spectroscopic scans. Centralizing or de-skewing such data is expensive in communication and often not permitted \cite{koutsoubis2024privacy}, and federated learning (FL) avoids centralizing it \cite{mcmahan2017communication}. Each client trains on its own compute, typically GPU nodes of its HPC system or nodes attached to the instrument, and the clients exchange only model updates over wide-area networks. The specialization of the instruments, however, makes the federated data strongly non-independent and identically distributed (non-IID) and long-tailed, which biases the global updates. Dominant clients overshadow rare but scientifically critical events. The volume of these data also makes training costly: in synchronous FedAvg (federated averaging, in which the server averages client models weighted by their data size), every round waits for all sampled clients, so the largest local datasets set the round time, the client computation, and the time until the global model reaches a target accuracy.
\begin{figure}[htbp]
\centering
\includegraphics[width=0.8\linewidth]{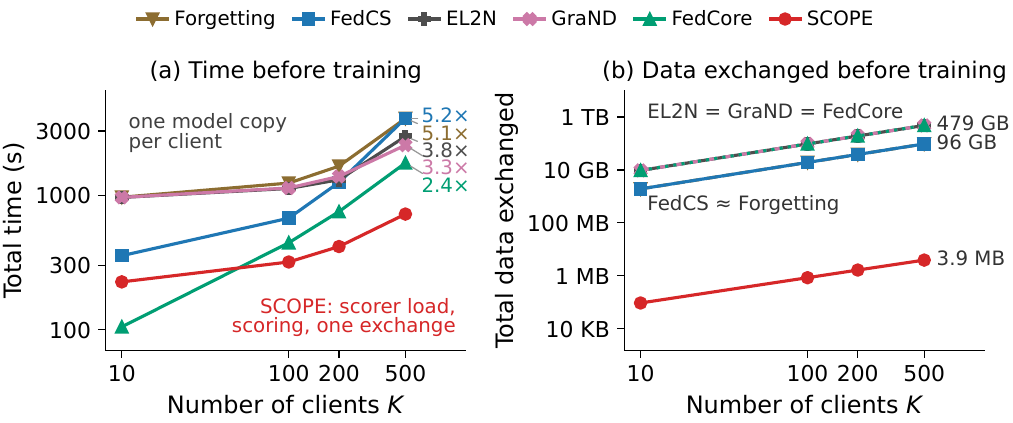}
\caption{Scaling with the number of clients $K$ on Tiny-ImageNet, total data fixed. (a) Time before training of all clients: selection, warm-up and one scorer load per client included, plus transfer at 200~Gb/s; labels give the ratio to SCOPE at $K = 500$. FedCore uses the initial model and is faster only at 10 clients. (b) Data exchanged before training: SCOPE sends a few statistics per class, the baselines full models; FedCore exchanges nothing.}
\label{fig:scaling}
\end{figure}
Coreset selection discards redundant samples while preserving accuracy \cite{marion2023less, patra2023calibrating}, which shortens every round, but it is difficult to deploy in FL. Clients that rely only on local heuristics make myopic choices and discard samples that are locally redundant but globally rare. Some federated methods assume a centralized proxy dataset for gradient matching \cite{sivasubramanian2024gradient}, which the privacy constraints of scientific collaborations rule out. Methods that rank samples by loss or gradient magnitude \cite{toneva2018empirical, paul2021deep} come from centralized training; scientific images often contain sensor noise and artifacts \cite{laine2021imaging, roth2021data}, which such scores can mistake for informative edge cases. Because clients also differ in labels and domains \cite{wang2025fpl}, local observations do not reflect the global data distribution, and a global view is needed. 

Coreset selection must not become the bottleneck it is meant to remove, but existing federated coreset methods score samples with the model being trained. FedCS pre-trains the model on every client before selecting \cite{hao2025fedcs}, and FedCore scores samples with gradients of the training model \cite{guo2024fedcore}. Their selection cost therefore grows with the size of the model, uses GPU time at every client before useful training begins, and must be repeated for every architecture; in our measurements, it also grows with the number of clients (Fig.~\ref{fig:scaling}). FedCS uploads one centroid of $D$-dimensional features per class, which gives $O(K \times C \times D)$ values for $K$ clients and $C$ classes \cite{hao2025fedcs}. Its traffic thus grows with both the federation and the model. Because selection happens only once, its value must be judged end to end, by the computation and communication that selection and training together need to reach a target accuracy.

\begin{figure*}[htbp]
    \centering
    \includegraphics[width=\linewidth]{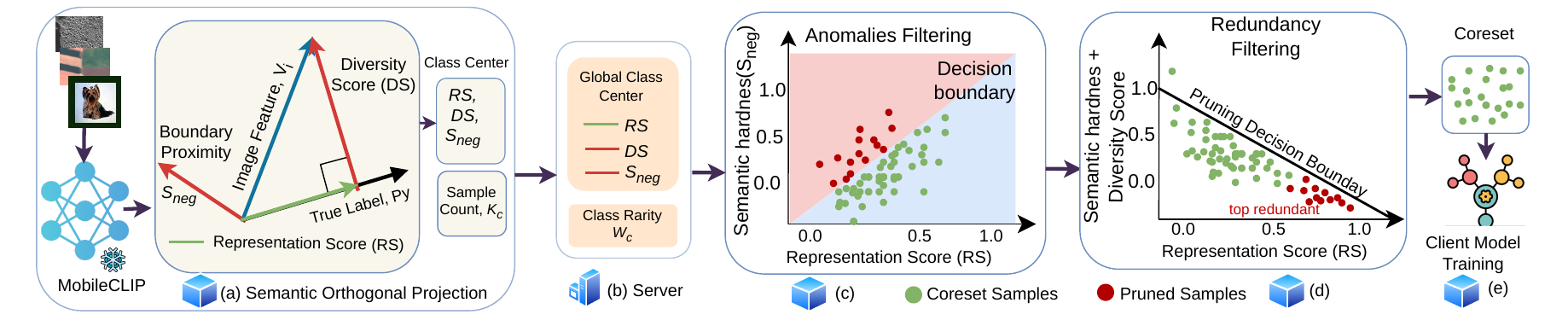}
    \caption{Overview of SCOPE. (a) Each client computes the scores $RS$, $DS$, and $S_{neg}$ of its samples with a frozen MobileCLIP-S2 model and uploads per-class statistics. (b) The server merges them into a global profile. (c) The Consensus Filter removes semantic anomalies. (d) The Balancing Filter removes redundant samples of classes that are locally abundant and globally common. (e) The remaining samples form the local coreset used in federated training.}
    \label{fig:scope_diagram}
\end{figure*}

We introduce SCOPE (Semantic Coreset using Orthogonal Projection Embeddings), a scalable, training-free federated coreset framework in which clients prune with a global view of the data at a small cost that does not depend on the model being trained. SCOPE scores the data once, before training, with a frozen, compact vision-language model (VLM) and exchanges only $O(K \times C)$ scalars with the server (Table~\ref{tab:intro_compare}, Fig.~\ref{fig:scope_diagram}). It thus removes the warm-up and the centroid upload, while still adopting the FedAvg paradigm. Its design addresses three questions.

\textbf{RQ1:} \textit{How can each client measure the utility of its samples without training, at a cost that does not depend on the model being trained?} SCOPE embeds each sample once with a frozen vision-language model and compares it with textual class prototypes, which gives a representation score, a diversity score derived from it, and a boundary proximity score (Section~\ref{subsec:metric_design}). These scores measure semantic utility rather than importance to the training model.

\textbf{RQ2:} \textit{How can the server build a global view of the data with an upload that stays small as the model and the number of clients grow?} Clients send per-class counts, means, and standard deviations of their three scores, which the server combines exactly into global statistics. The upload thus depends on the number of classes, not on the embedding dimension, and the statistics can be merged hierarchically or updated as new data arrive.

\textbf{RQ3:} \textit{How can clients prune without worsening global class imbalance?} Guided by the global profile and a server-side rarity weight, each client removes anomalies and then prunes redundant samples only in classes that are locally abundant and globally common, which protects classes that are locally frequent but globally rare. Our contributions are:
\begin{itemize}
    \item Training-free selection at each client (RQ1): each client scores its data with one forward pass of a frozen vision-language model, so selection needs no warm-up, its cost does not depend on the training model, and one coreset serves every architecture, with up to $34\times$ fewer FLOPs than baselines.

    \item A scalable, communication-light global profile (RQ2): each client uploads seven scalars per class, independent of the model, so the selection payload is up to $293\times$ smaller than the feature centroids of baselines; the server merges these statistics exactly and hierarchically, and SCOPE's total time before training grows less than that of the model-based baselines, which take 1.7 to $3.6\times$ longer at 500 clients.
    
    \item Pruning that preserves the global long tail (RQ3): clients prune only classes that are locally abundant and globally common; removing the global profile costs up to 24 points, and the server-side rarity weight guards the tail when targeting is broad; a conditional error bound shows when FedAvg trained on the coreset approaches the clean full-data objective.
    
    \item An evaluation that separates local computation, communication, and model performance, with communication converted to time at the 200~Gb/s line rate of the NCSA Delta network: on four datasets, including scientific micrographs, and against six baselines, SCOPE's coreset is the most accurate in 19 of 24 settings and reaches 98\% of the full-data accuracy in 18 to 42\% fewer global rounds than full-data training, with 8 to 21\% less client computation.
\end{itemize}
\begin{table}[htbp]
\centering
\caption{Coreset selection before training on Tiny-ImageNet with ResNet-50. Warm-up is the number of FedAvg rounds before selection, and Time is the GPU time of selection, warm-up included, of 100 clients with imbalance ratio $IR = 5$. Scaling is the time before training, transfer included, for 10 and 500 clients with the total data fixed at $IR = 2$ (Fig.~\ref{fig:scaling}a). SCOPE's times include one scorer load per client. FedCore's Scaling uses the initial model, its Warm-up and Time the model after five warm-up rounds. Reuse means that one coreset serves every training model.}
\label{tab:intro_compare}
\setlength{\tabcolsep}{2pt}
\begin{tabular}{lcccc}
\toprule
Method & Warm-up & Time & Scaling, $K{=}10 \to 500$ & Reuse \\
\midrule
EL2N        & 5 & 797 s      & 962 $\to$ 2{,}736 s      & No \\
GraND       & 5 & 840 s      & 963 $\to$ 2{,}355 s      & No \\
Forgetting  & 1 & 920 s      & 966 $\to$ 3{,}721 s      & No \\
FedCore     & 5 & 1{,}078 s  & 105 $\to$ 1{,}736 s      & No \\
FedCS       & 1 & 583 s      & 355 $\to$ 3{,}729 s      & No \\
SCOPE       & 0 & 321 s      & 233 $\to$ 1{,}049 s      & Yes \\
\bottomrule
\end{tabular}
\end{table}
\section{Related Work}
\par{Federated learning:}
FL systems reduce communication by compression and asynchrony \cite{lin2017deep, abrahamyan2021learned, chen2018adacomp, basu2019qsparse, xie2019asynchronous, ma2024fedstaleweight, wang2022asyncfeded} and computation by model compression \cite{jiang2022model, gao2024device, meng2023enhancing, jeong2018communication}, but treat the data as fixed. Federated coreset methods reduce the data itself: GCFL \cite{sivasubramanian2024gradient} needs a server-side dataset, FedCS \cite{hao2025fedcs} aggregates class feature centroids after warm-up training, and FedCore \cite{guo2024fedcore} builds per-client coresets from gradients of the training model; pruning heuristics designed for centralized training \cite{patra2023calibrating} ignore the global imbalance of a federation.
\par{Coreset selection:}
Forgetting Events \cite{toneva2018empirical}, GraND, and EL2N \cite{paul2021deep} keep samples with high loss or gradient norm; later work adds dynamic uncertainty \cite{he2024large}, training dynamics \cite{zhang2024spanning, qin2023infobatch}, and coreset selection for efficient quantization-aware training \cite{huang2023efficient}. On imbalanced data these signals also favor rare extreme cases \cite{hao2025fedcs}, and geometric methods such as Herding \cite{welling2009herding} and Moderate selection \cite{xia2022moderate} rely on Euclidean feature distances.
\par{Long-tailed FL and vision-language models:}
Federated long-tailed methods rebalance training itself, for example by re-training the classifier on federated features (CReFF \cite{shang2022creff}) or by calibrating local and generic heads (FedLoGe \cite{xiao2024fedloge}), and CLIP2FL \cite{shi2024clip2fl} distills a CLIP model into federated training on heterogeneous, long-tailed data. These methods change how the model is trained; SCOPE instead decides once, before training, which data each client trains on, so it is complementary to them and can precede other training methods, as our FedProx runs suggest (Table~\ref{tab:robust}).
\section{Problem Formulation}
A federation of $K$ clients, such as nodes or instruments, trains a model $w$ with synchronous FedAvg, in which a subset $\mathcal{S}_r$ of clients participates in round $r$. Client $k$ holds a private dataset $D_k$ with $n_{k,c}$ samples of class $c$, and $N_c = \sum_k n_{k,c}$ is the global count of class $c$. Two kinds of heterogeneity occur together. Globally, the class distribution is long-tailed: rare (tail) classes have far fewer samples than common (head) classes. Locally, the label mix differs sharply between clients, which we model with a Dirichlet distribution $\text{Dir}(\alpha)$. A class can therefore be abundant at one client and rare in the federation, which no client can tell from its own data.
\par{Selection problem:}
Before training, each client replaces $D_k$ with a coreset $\tilde{D}_k \subseteq D_k$. If client $k$ needs $\tau_k$ seconds of local computation per sample in a round and a round spends $t_{comm}$ on communication, round $r$ lasts
\begin{equation}
t_r = \max_{k \in \mathcal{S}_r} \tau_k \, |\tilde{D}_k| + t_{comm},
\qquad
\text{TTA} = t_{sel} + \sum_{r=1}^{R} t_r ,
\label{eq:tta}
\end{equation}
where $t_{sel}$ is the time of selection, including any warm-up and the upload of the selection statistics, and $R$ is the number of rounds until the global model reaches the target accuracy. We consider clients that are GPU nodes of the same type, so $\tau_k$ is equal across clients; in our evaluation, $t_{comm}$ is set by the 200~Gb/s network of the NCSA Delta cluster, a best case for communication, since a slower wide-area link would make the baselines' model uploads even costlier. The goal is the coreset $\{\tilde{D}_k\}$ that minimizes TTA for a given target accuracy; SCOPE does not solve this problem exactly but is designed around the requirements below. Pruning shortens each round, but removing the wrong samples raises $R$ or lowers the final accuracy, and a costly selection can outweigh the per-round savings. A useful selection therefore has to (i)~use global class information, so that pruning does not deepen the long tail; (ii)~cost little, with a time and an upload that do not grow with the model size $|w|$ or the feature dimension $D$; (iii)~not depend on the training model, so one coreset serves any architecture; and (iv)~exchange only aggregate statistics, never samples or embeddings.

\section{Methodology}
\subsection{System Overview}
SCOPE selects a coreset before federated training begins, without client-side training. Each client embeds its data once with MobileCLIP-S2, computes the $RS$, $DS$, and $S_{neg}$ scores defined in Section~\ref{subsec:metric_design}, and uploads per-class statistics of these scores. The server combines these into a global class profile and broadcasts it; no model updates or embeddings are exchanged. Each client then prunes in two stages (Section~\ref{subsec:local_coreset}): the Consensus Filter removes semantic anomalies, and the Balancing Filter removes redundant samples of classes that are locally abundant and globally common. Algorithm~\ref{alg:SCOPE_algo} summarizes the three phases.
\subsection{Semantic Metric Design in the Vision--Language Space}
\label{subsec:metric_design}
We formulate three data curation questions and answer each with a geometric score in the shared vision-language space. Let $v_i$ be the $L_2$-normalized image embedding of sample $i$ and $t_c$ the textual prototype of class $c$ \cite{radford2021learning}, the $L_2$-normalized mean of the text embeddings of five prompts, ``a photo of a \{c\}.'', ``a clear image of a \{c\}.'', ``a close-up of a \{c\}.'', ``a representation of the \{c\}.'', and ``a good photo of the \{c\}.'', where \{c\} is the class name; on Tiny-ImageNet, whose labels are WordNet identifiers, \{c\} is a neutral class token (``class 1'', ``class 2'', \ldots). For a sample with label $c_i$, the three scores are
\begin{equation}
\begin{aligned}
RS_i &= v_i \cdot t_{c_i}, \qquad DS_i = \| v_i - RS_i\, t_{c_i} \|_2 = \sqrt{1 - RS_i^2}, \\
S_{neg,i} &= \max_{j \neq c_i} v_i \cdot t_j .
\end{aligned}
\label{eq:scores}
\end{equation}
\subsubsection{Representation Score (RS):}
\textit{``How well does a sample represent its class?''} Samples that represent their class well anchor the coreset with the core features of each class and secure baseline accuracy. $RS_i$ is the cosine similarity between a sample and its class prototype, and a high value marks a textbook example. Rather than thresholding this scalar to discard data \cite{gadre2023datacomp}, we use it as the anchor for the other two scores.
\subsubsection{Diversity Score (DS), a reweighting of RS:}
\textit{``Does the sample contain structure beyond its class definition?''} A coreset of textbook examples alone would miss the visual variation within a class. We therefore isolate the part of the image embedding orthogonal to the class prototype, the rejection vector $v_i - RS_i\, t_{c_i}$, and take its length: a high $DS_i$ marks a sample with much structure outside its prototype, and a low value a prototypical one. Because both vectors are normalized, $DS_i$ is a deterministic function of $RS_i$ and carries no information beyond it. Its effect comes from the per-class standardization in Section~\ref{subsec:local_coreset}, where $DS$ has its own mean and variance; to first order $\hat{Z}_{DS,i} \approx 1 - \hat{Z}_{RS,i}$, so using $DS$ roughly doubles the weight of $RS$ relative to $S_{neg}$. Section~\ref{subsec:ablation} tests this reweighting.
\subsubsection{Boundary Proximity ($S_{neg}$) as decision-boundary ambiguity:}
\textit{``How easily could the sample be confused with another class?''} Samples near a decision boundary help the model learn where one class ends and the next begins. Following out-of-distribution scoring \cite{ming2022delving}, $S_{neg,i}$ is the largest similarity to any incorrect textual prototype, so a high value marks a sample near a decision boundary. Every client computes all prototypes itself from the known class names, so $S_{neg}$ is defined even for a client that holds a single class. All three scores cost one image-encoder pass per sample against prototypes computed once.
\subsection{Global Profile at the Server}
\par{Global Class Rarity:}
The server sums the class counts $n_{k,c}$ of all clients into $N_c$ and computes the global frequency $\pi_c = N_c / \sum_{c'} N_{c'}$ and the rarity weight $W_c \propto (\pi_c + \epsilon_0)^{-\gamma}$, with exponent $\gamma$ and stability constant $\epsilon_0$. A low $W_c$ marks a common class that may be pruned, and a high $W_c$ protects a rare class.
\par{Global Heterogeneity Profile:}
For each metric $m \in \{RS, DS, S_{neg}\}$ and class $c$, the server combines the client means $\mu_{m,c}^k$ and population standard deviations $\sigma_{m,c}^k$ into the sample-weighted global mean $\mu_{m,c}$ and, by the law of total variance, the global variance
\begin{equation}
\sigma_{m,c}^2 = \frac{1}{N_c} \sum_{k=1}^{K} n_{k,c} \Big[ \big(\sigma_{m,c}^k\big)^2 + \big(\mu_{m,c}^k - \mu_{m,c}\big)^2 \Big],
\label{eq:global_var}
\end{equation}
which accounts for both the spread within each client and the differences between clients. The profile is also mergeable: counts, means, and variances of disjoint sets combine exactly through the same decomposition, in any order and grouping \cite{chan1983algorithms}. The server can therefore aggregate hierarchically, for example per node and then globally, and a client that acquires new data only scores and uploads the statistics of the new samples.
\begin{figure}[htbp]
  \centering
\includegraphics[width=0.65\linewidth]{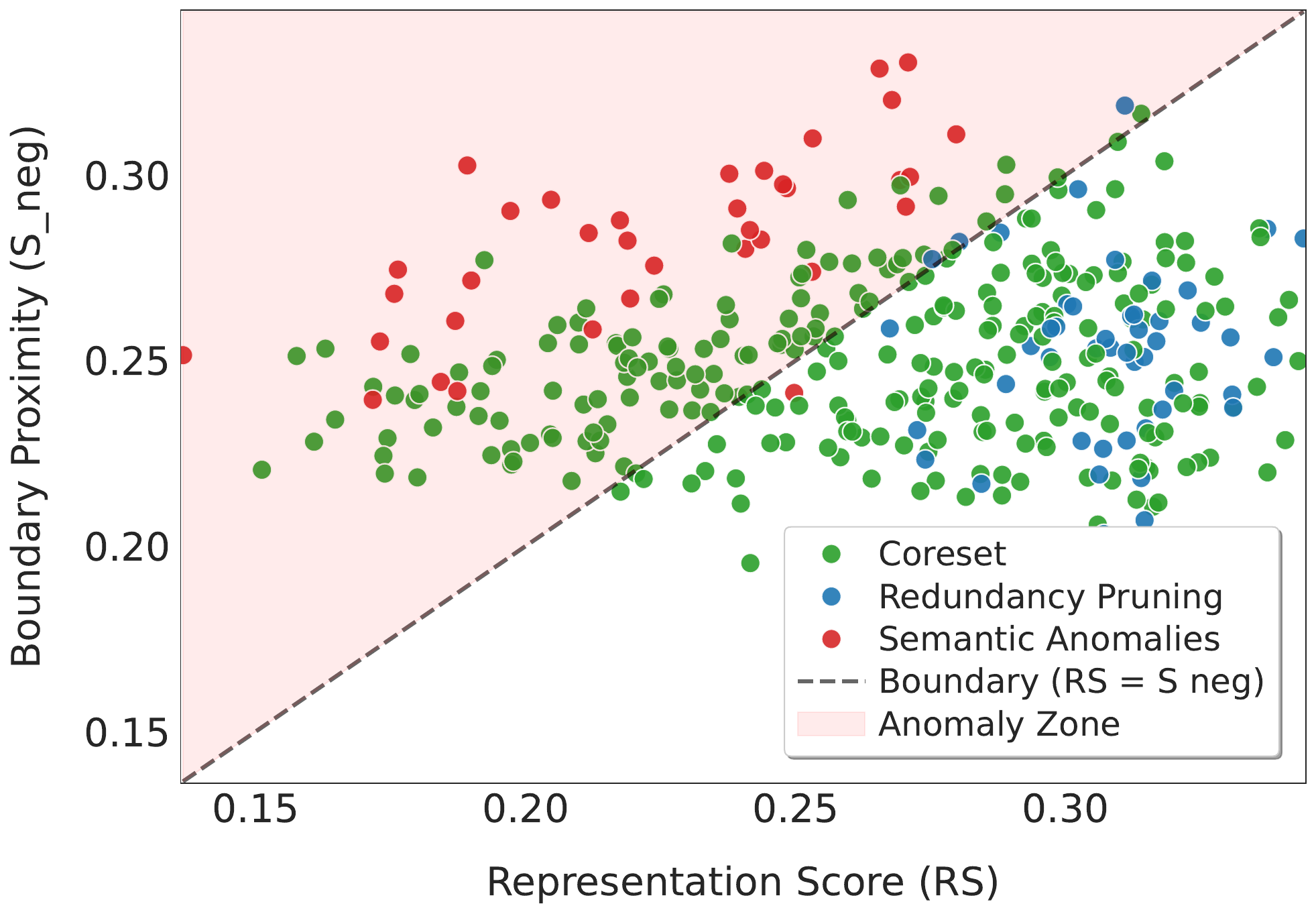}
\caption{Samples removed by the two filters on one client (CIFAR-100, $p_f = 0.5$, $p_l = 0.1$), plotted by representation score (RS) against boundary proximity ($S_{\text{neg}}$). In the shaded Anomaly Zone above the dashed line $RS = S_{\text{neg}}$, a sample is closer to another class than to its own. The axes show raw cosine similarities; the Consensus Filter ranks samples by the standardized score of Eq.~\eqref{eq:selection_scores}.}
\label{fig:selection_figure}
\end{figure}
\subsection{Local Coreset Selection}
\label{subsec:local_coreset}
Each client evaluates its local scores against the global profile in two stages; Fig.~\ref{fig:selection_figure} shows which samples the two filters remove. Because $RS$ and $S_{neg}$ are cosine similarities while $DS$ is a vector norm, the scores are first put on a common scale: each score $m_i$ of a sample of class $c$ is standardized with the global statistics, $Z_{m,i} = (m_i - \mu_{m,c}) / \sigma_{m,c}$, and mapped to $[0, 1]$ with $3\sigma$ bounds, $\hat{Z}_{m,i} = \min\big(1, \max(0, (Z_{m,i} + 3)/6)\big)$ (with $10^{-9}$ added to $\sigma_{m,c}$). The Consensus Filter removes $\lfloor p_l |D_k| \rfloor$ samples, the Balancing Filter keeps $\lfloor (1 - p_f) n'_{k,c} \rfloor$ samples of each target class, where $n'_{k,c}$ counts class $c$ after Stage~1, but at least one, ties keep the original sample order, and a class that a client does not hold has $f_c = 0$ and is never targeted. The two stages then rank samples by the Anomaly Score $AS_i$ and the Redundancy Score $R_i$:
\begin{equation}
\begin{aligned}
AS_i &= \hat{Z}_{S_{neg},i} - \hat{Z}_{RS,i}, \\
R_i &= \hat{Z}_{RS,i} - \hat{Z}_{S_{neg},i} - \hat{Z}_{DS,i}.
\end{aligned}
\label{eq:selection_scores}
\end{equation}
\par{Stage 1, Consensus Filter:}
\textit{``How much stronger is a sample's pull toward another class than toward its own?''} A large $AS_i$ marks a sample that resembles another class more than its own, typically a mislabeled or corrupted one. The client removes the $p_l$ fraction with the largest $AS_i$, which leaves the clean set $D'_k$.
\par{Stage 2, Balancing Filter:}
\textit{``Does a sample add information, or does it duplicate the class core?''} A high $R_i$ marks a typical sample that is neither diverse nor close to another class, and hence redundant. To decide which classes are pruned, the client computes the Targeting Metric $T_c = f_c / W_c$, where $f_c$ is the share of class $c$ in $D'_k$. A high $T_c$ flags a class that is locally abundant and globally common, and the target classes are those within a relative margin $\beta$ of the largest value, $(\max_{c'} T_{c'} - T_c) / (\max_{c'} T_{c'} + \epsilon_0) \le \beta$. In every target class, the client removes the top $p_f$ fraction of samples by $R_i$, which discards prototypical samples and keeps the diverse samples near the class boundaries. The remaining data form the local coreset $\tilde{D}_k$. After selection, SCOPE and the baselines train on their coresets in the same way, with the same synchronous FedAvg rounds and hyperparameters, so the methods differ only in the coreset and in the cost of selecting it, which Sections~\ref{subsec:cost} and~\ref{subsec:comm} measure.
\begin{algorithm}[!t]
\caption{SCOPE: Federated Semantic Coreset Selection}
\label{alg:SCOPE_algo}
\small
\begin{algorithmic}[1]
\Require Local datasets $\{D_k\}$, class names, rates $p_l, p_f$, margin $\beta$, exponent $\gamma$
\Ensure Local coresets $\{\tilde{D}_k\}$
\Statex \textbf{Phase 1: Client-side scores}
\For{each client $k$ \textbf{in parallel}}
    \State Compute prototypes $t_c$; for each $x_i \in D_k$, compute $RS_i$, $DS_i$, $S_{neg,i}$ by Eq.~\eqref{eq:scores}
    \State Upload $n_{k,c}$, $\mu_{m,c}^k$, $\sigma_{m,c}^k$ for each local class $c$ and $m \in \{RS, DS, S_{neg}\}$
\EndFor
\Statex \textbf{Phase 2: Server-side global profile}
\State For each class $c$: $N_c \gets \sum_{k} n_{k,c}$, $W_c \propto (N_c / \sum_{c'} N_{c'} + \epsilon_0)^{-\gamma}$, and $\mu_{m,c}$, $\sigma_{m,c}$ by Eq.~\eqref{eq:global_var}
\State Broadcast $\Omega = \{W_c, \mu_{m,c}, \sigma_{m,c}\}$ to all clients
\Statex \textbf{Phase 3: Client-side coreset selection}
\For{each client $k$ \textbf{in parallel}}
    \State Standardize the scores with $\Omega$; compute $AS_i$ and $R_i$ by Eq.~\eqref{eq:selection_scores}
    \State $D'_k \gets D_k$ without the top $p_l$ fraction by $AS_i$ \Comment{anomalies}
    \State $T_c \gets f_c / W_c$, with $f_c$ the share of class $c$ in $D'_k$
    \State $\mathcal{C}_{target} \gets \{ c \mid (\max_{c'} T_{c'} - T_c)/(\max_{c'} T_{c'} + \epsilon_0) \le \beta \}$
    \State $\tilde{D}_k \gets D'_k$ without the top $p_f$ fraction of each class in $\mathcal{C}_{target}$ by $R_i$, keeping at least one \Comment{redundancy}
\EndFor
\end{algorithmic}
\end{algorithm}
\section{Theoretical Analysis}
\label{sec:convergence}
This section states what the two filters must achieve for FedAvg on the coreset to approach a stationary point of the clean objective. The statements are conditional: their assumptions are not implied by the method, and Section~\ref{subsec:theory_validation} tests their consequences on two datasets. They do not by themselves explain SCOPE's accuracy.
\subsection{Notation and Assumptions}
Client $k$ has raw (possibly noisy) data $D_k$, correctly labeled samples $D_k^{clean} \subset D_k$, and coreset $\tilde{D}_k \subset D_k$, with empirical distributions $P_k$, $P_k^{clean}$, and $\widetilde{P}_k$. For a per-sample loss $\ell(w;\xi)$, the local objectives are the means $F_k$, $F_k^{clean}$, and $\widetilde{F}_k$ of $\ell$ under these distributions. The clean objective $F = \sum_k p_k F_k^{clean}$ uses target weights $p_k$, and the coreset objective $\widetilde{F} = \sum_k q_k \widetilde{F}_k$ uses the weights $q_k$ of the objective that FedAvg actually optimizes; both weight vectors are nonnegative and sum to one. With $E$ local epochs and data-size aggregation, $q_k$ is proportional to $|\tilde{D}_k|$ times the number of local steps \cite{wang2020tackling}, so selection changes the weights. Let $P = \sum_k p_k P_k$, $\widetilde{P} = \sum_k q_k \widetilde{P}_k$, and $F^{raw} = \sum_k p_k F_k$. Client drift on the raw data and on the coreset is $\Gamma = \sup_w \sum_k p_k \| \nabla F_k - \nabla F^{raw} \|^2$ and $\widetilde{\Gamma} = \sup_w \sum_k q_k \| \nabla \widetilde{F}_k - \nabla \widetilde{F} \|^2$, and $G = \sup_{w,k} \| \nabla F_k^{clean}[w] - \nabla F[w] \|$ is the largest dissimilarity of the clean client objectives, assumed finite.
\begin{assumption}[Semantic Gradient Deviation]
\label{ass:semantic}
Let $a_i = (1 + AS_i)/2 \in [0, 1]$, which ranks samples as $AS_i$ does. There are $L_{sem}, \delta \ge 0$ such that $\| \nabla \ell(w;\xi_i) - \nabla F_k^{clean}[w] \| \le L_{sem} a_i + \delta$ for every $w$, client $k$, and $\xi_i \in D_k$.
\end{assumption}
\begin{assumption}[Gradient Lipschitz in the Data]
\label{ass:data_lipschitz}
For a metric $d$ on samples, $\| \nabla \ell(w;\xi) - \nabla \ell(w;\xi') \| \le L_x \, d(\xi, \xi')$ for all $w$, $\xi$, and $\xi'$.
\end{assumption}
Both assumptions are strong. Assumption~\ref{ass:semantic} asks a frozen VLM score to bound the gradient deviation of an arbitrary training model, and in Assumption~\ref{ass:data_lipschitz}, $d$ must account for labels, for example by adding a label-mismatch penalty to the distance between VLM embeddings.
\subsection{Gradient Bias and Client Drift of the Coreset}
\begin{lemma}[Gradient Bias of the Coreset]
\label{lemma:bias_reduction}
Let $\bar{a}_k$ be the largest $a_i$ among the samples kept on client $k$. Under Assumption~\ref{ass:semantic}, $\epsilon_k = \sup_w \| \nabla \widetilde{F}_k[w] - \nabla F_k^{clean}[w] \| \le L_{sem} \bar{a}_k + \delta$, and for every $w$,
\begin{equation}
\| \nabla \widetilde{F}[w] - \nabla F[w] \| \le \epsilon = \textstyle\sum_k q_k (L_{sem} \bar{a}_k + \delta) + 2\, d_{TV}(q, p)\, G,
\label{eq:bias}
\end{equation}
where $d_{TV}(q, p) = \frac{1}{2} \sum_k |q_k - p_k|$.
\end{lemma}
\begin{proof}
$\nabla \widetilde{F}_k - \nabla F_k^{clean}$ is the mean of $\nabla \ell(w;\xi_j) - \nabla F_k^{clean}$ over the kept samples, so its norm is at most the largest of these norms, which Assumption~\ref{ass:semantic} bounds by $L_{sem} \bar{a}_k + \delta$. Next, $\nabla \widetilde{F} - \nabla F = \sum_k q_k (\nabla \widetilde{F}_k - \nabla F_k^{clean}) + \sum_k (q_k - p_k)(\nabla F_k^{clean} - \nabla F)$, where we used $\sum_k (q_k - p_k) = 0$; the triangle inequality bounds the two sums by the two terms of Eq.~\eqref{eq:bias}.
\end{proof}
The first term shrinks when Stage~1 removes the samples with the largest $a_i$; the second is the cost of changing the client weights, which vanishes when selection keeps the weights unchanged. The coreset of client $k$ is less biased than its raw data, whose bias is $\zeta_{noise} = \sup_w \| \nabla F_k - \nabla F_k^{clean} \|$, if $L_{sem} \bar{a}_k + \delta < \zeta_{noise}$.
\begin{lemma}[Client Drift and Transport Distance]
\label{lemma:drift_reduction}
Let $W_1$ be the 1-Wasserstein distance under the metric $d$. Under Assumption~\ref{ass:data_lipschitz}, $\Gamma \le L_x^2 \sum_k p_k W_1(P_k, P)^2$ and $\widetilde{\Gamma} \le L_x^2 \sum_k q_k W_1(\widetilde{P}_k, \widetilde{P})^2$. If $W_1(\widetilde{P}_k, \widetilde{P}) \le \rho \, W_1(P_k, P)$ for every client, then $\widetilde{\Gamma} \le \rho^2 L_x^2 \sum_k q_k W_1(P_k, P)^2$.
\end{lemma}
\begin{proof}
Fix $w$, $k$, and a unit vector $u$. By Assumption~\ref{ass:data_lipschitz}, $\xi \mapsto \langle u, \nabla \ell(w;\xi) \rangle / L_x$ is 1-Lipschitz under $d$. Since $\nabla F_k = \mathbb{E}_{P_k} \nabla \ell$ and $\nabla F^{raw} = \mathbb{E}_{P} \nabla \ell$, the Kantorovich--Rubinstein duality \cite{villani2009optimal} gives $\langle u, \nabla F_k - \nabla F^{raw} \rangle \le L_x W_1(P_k, P)$; choosing $u$ along $\nabla F_k - \nabla F^{raw}$, squaring, weighting, summing, and taking the supremum over $w$ gives the first bound. The same steps give the second, and the contraction gives the third.
\end{proof}
\subsection{Conditional Convergence}
\begin{proposition}[Stationarity Transfer]
\label{prop:transfer}
Under Assumption~\ref{ass:semantic}, every $w$ satisfies $\| \nabla F[w] \|^2 \le 2 \| \nabla \widetilde{F}[w] \|^2 + 2 \epsilon^2$. Hence, if FedAvg on the coreset returns $\hat{w}$ with $\mathbb{E} \| \nabla \widetilde{F}[\hat{w}] \|^2 \le \nu$, then $\mathbb{E} \| \nabla F[\hat{w}] \|^2 \le 2\nu + 2\epsilon^2$.
\end{proposition}
\begin{proof}
By Lemma~\ref{lemma:bias_reduction}, $\| \nabla F[w] \| \le \| \nabla \widetilde{F}[w] \| + \epsilon$, and $(a + b)^2 \le 2a^2 + 2b^2$; take expectations for the second claim.
\end{proof}
For $L$-smooth losses with bounded gradient variance, nonconvex FedAvg analyses provide such a $\nu$ for the objective FedAvg optimizes \cite{karimireddy2020scaffold,wang2020tackling}: it decreases with the number of rounds and grows with the drift $\widetilde{\Gamma}$ and with partial participation. The proposition thus separates two effects. The coreset's gradient bias sets an error floor $2\epsilon^2$ that more rounds do not remove, and client drift, which Lemma~\ref{lemma:drift_reduction} bounds by a transport distance, affects how quickly FedAvg reaches it.
\begin{table*}[htbp]
\centering
\caption{Cost of coreset selection and of training to a target accuracy on Tiny-ImageNet  ($IR = 5$, $p_f = 0.5$), with 100 clients and 10 sampled per round. The left part gives the selection cost of all clients before training. The right part gives the rounds, GPU time, upload, and transfer time needed to reach 95 and 98\% of the Full DB accuracy, with selection included. Transfer times use the 200~Gb/s line rate of the Delta network without latency, so they are lower bounds. SCOPE's compute includes one scorer load per client; its energy does not. Bold marks the best coreset method; Random scores nothing and is not ranked in the selection columns; energy is from one seed.}
\label{tab:cost}
\footnotesize
\setlength{\tabcolsep}{2.2pt}
\resizebox{\textwidth}{!}{%
\begin{tabular}{l ccccccc c cccc}
\toprule
& \multicolumn{7}{c}{Coreset selection} & & \multicolumn{4}{c}{Training to 95 / 98\% of Full DB} \\
\cmidrule(lr){2-8} \cmidrule(lr){10-13}
Method & Ep & Compute (s) & Energy (kJ) & Exch. & Upload & Download & Transfer (s) & & Rounds & Compute (s) & Upload (GB) & Transfer (s) \\
\midrule
Full DB & -- & -- & -- & -- & -- & -- & -- & & 51 / 109 & 639 / 1{,}366 & 48.9 / 104.5 & 3.9 / 8.4 \\
\midrule
Random  & 0  & 67      & --    & 0 & 0       & 0       & 0    & & 83 / 147    & 900 / 1{,}542     & 79.6 / 140.9  & 6.4 / 11.3 \\
EL2N    & 25 & 797     & 102.6 & 5 & 47.9 GB & 47.9 GB & 3.84 & & 58.5 / 108  & 1{,}376 / 1{,}866 & 104.0 / 151.5 & 8.3 / 12.1 \\
GraND   & 25 & 840     & 106.1 & 5 & 47.9 GB & 47.9 GB & 3.84 & & 75.5 / 119  & 1{,}565 / 1{,}982 & 120.3 / 162.0 & 9.6 / 13.0 \\
FedCS   & 5  & 583     & 52.0  & 2 & 9.6 GB  & 9.6 GB  & 0.77 & & 57 / 106    & 1{,}145 / 1{,}627 & 64.3 / 111.2  & 5.1 / 8.9 \\
FedCore & 25 & 1{,}078 & 126.5 & 5 & 47.9 GB & 47.9 GB & 3.84 & & 56 / \textbf{76} & 1{,}627 / 1{,}823 & 101.6 / 120.8 & 8.1 / 9.7 \\
\midrule
SCOPE   & \textbf{0} & \textbf{321} & \textbf{28.9} & \textbf{1} & \textbf{0.27 MB} & \textbf{0.56 MB} & \textbf{$<$0.001} & & \textbf{53.5} / 83.5 & \textbf{853} / \textbf{1{,}150} & \textbf{51.3} / \textbf{80.1} & \textbf{4.1} / \textbf{6.4} \\
\bottomrule
\end{tabular}%
}
\end{table*}

\section{Experimental Evaluation}
We evaluate SCOPE in the order in which a federation uses it. Q1: What does coreset selection cost each client in local computation, and how does this cost grow with the training model and the number of clients (Section~\ref{subsec:cost})? Q2: How much does each method communicate, before training and until the global model reaches the target accuracy (Section~\ref{subsec:comm})? Q3: Does the coreset preserve accuracy, including on globally rare classes, and how many global rounds does training need (Section~\ref{subsec:accuracy})? Q4: How robust and general is the coreset, and do the filters behave as the analysis assumes (Sections~\ref{subsec:backbone} and~\ref{subsec:theory_validation})? Q1 and Q2 test RQ1 and RQ2, Q3 tests RQ3, and Q4 tests the generality.
\subsection{Experimental Setup}
\par{Datasets and long-tailed federations:}
\label{subsec:settings}
CIFAR-10 uses ResNet-18 on 10 clients with full participation; Tiny-ImageNet uses ResNet-50 and CIFAR-100 uses ViT-B-16, each on 100 clients with 10 per round; the Ultrahigh Carbon Steel Micrograph DataBase (UHCS) \cite{decost2017uhcsdb} uses Swin-Tiny on 10 clients. An imbalance ratio $IR \in \{2, 5, 10\}$, the ratio between the sample counts of the most and the least frequent class, makes the global distribution long-tailed. On top of that, we apply a Dirichlet partition Dir($\alpha$), $\alpha \in \{0.1, 1.0\}$. The original test splits are class-balanced except for UHCS (366 images, 3 to 225 per class), where one standard error near 93\% accuracy is about 1.3 points. We use $\beta = 0.5$, $\gamma = 0.7$, $\epsilon_0 = 10^{-8}$, and $p_l = 0.1$ for every dataset (Table~\ref{tab:beta_gamma}), without per-dataset tuning or a validation split.
\subsubsection{Baselines and federated training}
We compare with full-data FedAvg (Full DB) \cite{mcmahan2017communication}, Random, and five coreset methods that differ in how each client warms up before selecting the coreset. EL2N and GraND \cite{paul2021deep} rank samples by the error or gradient norm of the training model after 5 warm-up rounds of FedAvg on all local data. Forgetting \cite{toneva2018empirical} counts forgetting events while each client trains locally, followed by one aggregation. FedCore \cite{guo2024fedcore} ranks samples by last-layer gradients of the training model, using the initial model or, where stated, the model after 5 warm-up rounds. FedCS \cite{hao2025fedcs} pre-trains the model federatedly; each client computes per-class centroids of the model's features, and the server takes their per-class median as the global center. Our reproduction uses 5 local epochs, one FL aggregation, and the classifier output as features. SCOPE, in contrast, needs no warm-up, scores samples with a frozen VLM, and exchanges only per-class score statistics. All methods train for $T = 200$ rounds with SGD, momentum 0.9, no weight decay, batch size 64, $E = 5$ local epochs unless stated, and a cosine learning rate from 0.0125 to 0; all models except ResNet-18 start from ImageNet-pretrained weights. After coreset selection, every method restarts from the initial model, so the baselines' warm-up only serves to score samples. Results are averaged over two seeds unless stated. SCOPE scores each sample with one forward pass of the frozen MobileCLIP-S2 encoder.
\subsubsection{Accuracy and cost metrics}
\emph{Accuracy} is the mean top-1 test accuracy over the last 10 FL rounds, which equals the mean per-class accuracy on the balanced test splits. \emph{Client accuracy} is the accuracy of the global model on one client's label mix, and its mean over the 10\% worst clients shows which clients lose from pruning. \emph{Rounds to target} is the number of global rounds until the 5-round moving average of test accuracy first reaches 95 or 98\% of the Full DB accuracy. \emph{Selection computation} is the GPU time that all clients together spend on warm-up, scoring, and pruning before training. Loading the training model is common to all methods and not counted; SCOPE's scorer serves only selection, so its load is counted per client. \emph{Communication} is the number of bytes exchanged with the server, converted into transfer time at the 200~Gb/s line rate of the NCSA Delta network; these times are lower bounds. \emph{Training computation} is the GPU time of the sampled clients, added over the rounds until the target. In the multi-GPU measurement on Tiny-ImageNet with 100 clients, the clients are spread over 1, 4, or 8 GPUs, and each GPU scores its clients one after another (Table~\ref{tab:dist}). We run the clients on GPUs of the same type, so client computation is the measured GPU time summed over all clients.
\begin{figure*}[htbp]
\centering
\includegraphics[width=0.95\textwidth]{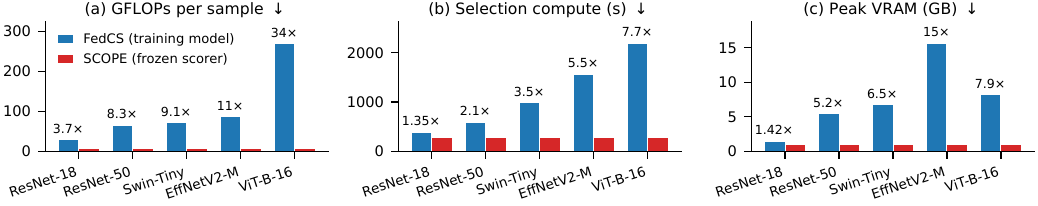}
\caption{Coreset selection cost for different training models on CIFAR-100. FedCS selects with the model being trained, so it must warm the model up and pass every sample through it, while SCOPE runs one pass of the same frozen MobileCLIP-S2 encoder for every model. (a) Computation per sample; (b) GPU time of selection, all clients; (c) peak GPU memory. Labels give FedCS's cost relative to SCOPE's. Lower is better in all panels.}
\label{fig:model_indep}
\end{figure*}
\begin{table*}[htbp]
\centering
\caption{Accuracy (\%) of the global model trained on each coreset with $p_l = 0.1$, at two of the three pruning rates per dataset. SCOPE is the best coreset method in 19 of all 24 settings. CIFAR-10 and UHCS use 10 clients; Tiny-ImageNet (TIN) and CIFAR-100 use 100 clients with 10 per round; $\alpha = 0.1$ unless stated. Bold marks the best coreset method and div.\ a diverged run; Forgetting on CIFAR-10 and Random on CIFAR-100 use one seed. UHCS accuracy moves in steps of 0.27.}
\label{tab:acc_main}
\scriptsize
\setlength{\tabcolsep}{2.5pt}
\begin{tabular}{l cc cc cc cc cc cc cc cc}
\toprule
& \multicolumn{2}{c}{CIFAR-10} & \multicolumn{2}{c}{CIFAR-10} & \multicolumn{2}{c}{UHCS} & \multicolumn{2}{c}{UHCS} & \multicolumn{2}{c}{TIN, $\alpha{=}1$} & \multicolumn{2}{c}{TIN} & \multicolumn{2}{c}{CIFAR-100} & \multicolumn{2}{c}{CIFAR-100} \\
& \multicolumn{2}{c}{$IR = 2$} & \multicolumn{2}{c}{$IR = 10$} & \multicolumn{2}{c}{$IR = 2$} & \multicolumn{2}{c}{$IR = 10$} & \multicolumn{2}{c}{$IR = 2$} & \multicolumn{2}{c}{$IR = 5$} & \multicolumn{2}{c}{$IR = 2$} & \multicolumn{2}{c}{$IR = 10$} \\
\cmidrule(lr){2-3} \cmidrule(lr){4-5} \cmidrule(lr){6-7} \cmidrule(lr){8-9} \cmidrule(lr){10-11} \cmidrule(lr){12-13} \cmidrule(lr){14-15} \cmidrule(lr){16-17}
$p_f$ & 0.1 & 0.9 & 0.1 & 0.9 & 0.1 & 0.5 & 0.1 & 0.5 & 0.5 & 0.9 & 0.5 & 0.9 & 0.1 & 0.5 & 0.1 & 0.5 \\
\midrule
Random     & 41.69 & 46.50 & 44.84 & 41.91 & 88.52 & 86.61 & 91.53 & 92.08 & 57.36 & 55.67 & 53.89 & 52.50 & 84.76 & 84.15 & 80.13 & 79.98 \\
Forgetting & 55.49 & 46.61 & 44.10 & div. & 84.43 & 86.34 & 91.26 & 89.89 & 58.15 & 56.57 & 54.21 & 54.04 & 84.51 & 83.76 & 80.55 & 80.30 \\
EL2N       & 47.27 & 42.19 & 37.54 & 40.58 & 85.25 & 81.15 & 92.62 & 88.52 & 59.13 & 58.18 & \textbf{54.34} & 53.12 & 83.80 & 82.56 & 75.93 & 76.81 \\
GraND      & 48.44 & 39.83 & 36.53 & 15.93 & 82.24 & 78.96 & 91.53 & 92.08 & 58.88 & 58.33 & 54.14 & 52.58 & 83.80 & 82.57 & 75.96 & 77.15 \\
FedCS      & 53.09 & 47.39 & 43.40 & 42.76 & 86.61 & 85.79 & 93.17 & 92.08 & 58.84 & 58.25 & 53.67 & 52.57 & 84.56 & 84.23 & 80.52 & 80.18 \\
FedCore    & 55.96 & 48.22 & 44.98 & 42.52 & 83.06 & 87.43 & 92.08 & 89.07 & 58.13 & 58.64 & 53.88 & 52.42 & 84.40 & 84.31 & 79.97 & 80.13 \\
\midrule
SCOPE      & \textbf{56.48} & \textbf{48.33} & \textbf{45.65} & \textbf{42.80} & \textbf{95.36} & \textbf{94.26} & \textbf{95.36} & \textbf{94.81} & \textbf{59.27} & \textbf{58.78} & 54.28 & \textbf{55.38} & \textbf{85.10} & \textbf{84.68} & \textbf{80.92} & \textbf{80.62} \\
\midrule
Full DB    & \multicolumn{2}{c}{55.63} & \multicolumn{2}{c}{45.07} & \multicolumn{2}{c}{87.70} & \multicolumn{2}{c}{93.99} & \multicolumn{2}{c}{59.85} & \multicolumn{2}{c}{54.41} & \multicolumn{2}{c}{85.09} & \multicolumn{2}{c}{85.94} \\
\bottomrule
\end{tabular}
\end{table*}
\subsubsection{Coreset selection budget}
$p_l$ is the fraction of each client's samples that the Consensus Filter removes as anomalies, and $p_f$ is the fraction of each target class that the Balancing Filter removes as redundant; classes outside the target set keep all their samples. Because $p_f$ applies only to the target classes, a client keeps much more than $1 - p_f$ of its data, for example 69\% at $p_f = 0.9$ (Table~\ref{tab:beta_gamma}).
\subsection{Local Computation of Coreset Selection}
\label{subsec:cost}
A SCOPE client passes each sample once through a frozen encoder and never trains. Every baseline in Table~\ref{tab:cost} scores samples with the training model, so its clients must first train that model on all their data, for one aggregation in FedCS and for five FedAvg rounds in EL2N, GraND, and FedCore. This warm-up is what makes them expensive: on Tiny-ImageNet with 100 clients, SCOPE's selection takes 321~s in total, including one scorer load per client, while FedCS needs 583~s and the methods with five warm-up rounds up to 1{,}078~s in Table~\ref{tab:cost}. Energy follows the same order, and only Random, which scores nothing, is cheaper.
\par{Cost independent of the training model:}
SCOPE's cost is set by its scorer, so every sample costs the same 7.9~GFLOPs whatever model is trained. FedCS selects with the training model, so its cost grows with that model: it needs up to $34\times$ more FLOPs per sample (Fig.~\ref{fig:model_indep}a), and its selection is $1.35\times$ slower than SCOPE's for ResNet-18 and $7.7\times$ slower for ViT-B-16 (Fig.~\ref{fig:model_indep}b). SCOPE also needs 1.4 to $15\times$ less peak GPU memory than FedCS for every training model in Fig.~\ref{fig:model_indep}c.
\par{Scaling with the number of clients:}
\label{subsec:scaling}
With the total data fixed, SCOPE's scoring work does not change with the number of clients, since each sample passes the encoder once. Each client adds only one scorer load of about 1.7~s, so SCOPE's time grows from 233 to 1{,}049~s between 10 and 500 clients. Each added client in a baseline instead trains or scores its own model copy and exchanges it, reaching up to $3.6\times$ SCOPE's time at 500 clients in Fig.~\ref{fig:scaling}a.
\subsection{Coreset Selection and Training: Communication Cost}
\label{subsec:comm}
\par{One exchange before training:}
Each SCOPE client uploads seven statistics per class and receives the global profile, 0.27~MB up and 0.56~MB down for 100 clients, which Delta moves in microseconds. A warm-up baseline sends the 95.9~MB model to the server and back every round, 9.6~GB each way per round and 47.9~GB for five (Table~\ref{tab:cost}). FedCS also needs a second exchange for class centroids. FedCore with the initial model and Random send nothing before training.
\par
\begin{figure}[htbp]
\centering
\includegraphics[width=0.65\linewidth]{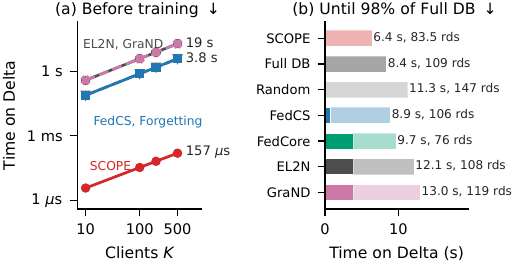}
\caption{Communication time on the 200~Gb/s Delta network, up and down, on Tiny-ImageNet, a lower bound without latency. (a) Before training, for the data of Fig.~\ref{fig:scaling}b. (b) Until 98\% of the Full DB accuracy; dark: coreset selection, light: training rounds. Lower is better.}
\label{fig:comm}
\end{figure}

{Payload and scaling with the number of clients:}
SCOPE's payload depends on the number of classes, not on the model. A centroid method sends a $D$-dimensional vector per class instead of seven values, which makes its upload $73\times$ to $293\times$ larger for our training models; in a separate FedCS run with 2048-dimensional penultimate features on Tiny-ImageNet, we measure $293\times$ at similar accuracy. Because every added client exchanges its own model, the traffic grows linearly with the federation (Fig.~\ref{fig:scaling}b): at 500 clients, five warm-up rounds exchange 479~GB, which occupy the Delta network for at least 19~s, against 3.9~MB and 157~$\mu$s for SCOPE (Fig.~\ref{fig:comm}a). On UHCS, which has six classes, the classifier-output centroids of FedCS have six values per class, slightly fewer than SCOPE's seven, but FedCS first uploads 1.1~GB of warm-up models.
\par{Traffic until the target accuracy:}
All methods then train the same model, so each round costs the same 1.92~GB, and the total traffic depends on the number of rounds. Because SCOPE needs fewer rounds and no warm-up, its communication until 98\% of the Full DB accuracy takes at least 6.4~s on Delta, selection included, against 8.4~s for Full DB and 8.9 to 13.0~s for the baselines that warm up (Fig.~\ref{fig:comm}b).
\subsection{Accuracy and Convergence of the FL Model}
\label{subsec:accuracy}
\label{subsec:e2e}
\par{Equal pruning rates:}
SCOPE is the best coreset method in 15 of the 16 columns of Table~\ref{tab:acc_main}, with the largest margins on the scientific micrographs of UHCS, where it leads the best baseline by 2 to 7 points. With light pruning, it stays within one point of Full DB or above it in most settings, because the samples removed first are redundant samples of common classes. SCOPE also keeps more data than the baselines at the same $p_f$, because it prunes only classes that are locally abundant and globally common.
\begin{figure*}[htbp]
\centering
\includegraphics[width=\textwidth]{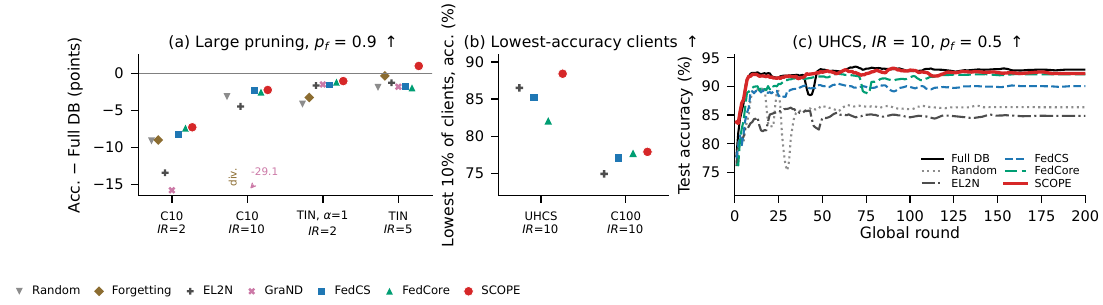}
\caption{Accuracy of the global model trained on each coreset. (a) Accuracy relative to Full DB under large pruning, $p_f = 0.9$; C10: CIFAR-10, TIN: Tiny-ImageNet, div.: diverged. (b) Mean accuracy of the 10\% of clients with the lowest accuracy at $p_f = 0.9$, one seed. (c) Test accuracy over global rounds on UHCS with $IR = 10$ and $p_f = 0.5$; FedCore uses five warm-up rounds. Higher is better in all panels.}
\label{fig:perf}
\end{figure*}
\begin{figure*}[htbp]
\centering
\includegraphics[width=\textwidth]{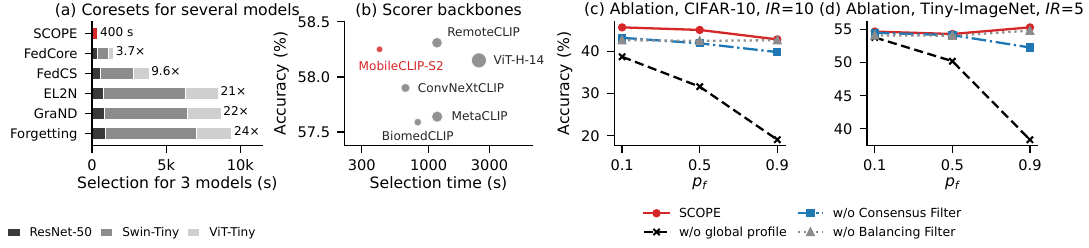}
\caption{(a) Selection time, one seed, to provide coresets for three training models on Tiny-ImageNet; each baseline selects again for every model, while SCOPE selects once, including one scorer load per client, and labels give the ratio to SCOPE. (b) Accuracy and selection time of different scorer backbones on Tiny-ImageNet; marker area shows peak GPU memory. (c, d) Accuracy on CIFAR-10 and Tiny-ImageNet when the global profile, the Consensus Filter, or the Balancing Filter is removed.}
\label{fig:generality}
\label{fig:ablation}
\end{figure*}
\par
\begin{figure}[htbp]
\centering
\includegraphics[width=0.50\linewidth]{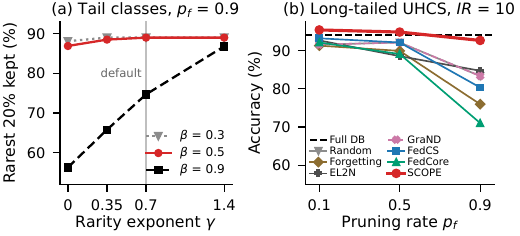}
\caption{Selection parameters and accuracy on a long-tailed federation. (a) Share of the 20\% rarest classes of Tiny-ImageNet that SCOPE keeps at $p_f = 0.9$, against the rarity exponent $\gamma$ for three targeting margins $\beta$. (b) Accuracy on UHCS with $IR = 10$ as the pruning rate grows.}
\label{fig:params}
\end{figure}

{Large pruning and strong imbalance:}
The advantage of a frozen scorer grows as more data is removed (Fig.~\ref{fig:perf}a). The baselines score samples with a warm-up model trained on skewed local data, so their scores follow the local skew, and they prune without knowing how rare a class is in the federation. At $p_f = 0.9$, GraND therefore loses 29 points on CIFAR-10 with $IR = 10$, where Forgetting diverges. SCOPE scores with a frozen encoder that never sees the local skew and is the coreset method closest to Full DB in all four settings; on Tiny-ImageNet with $IR = 5$ it even exceeds Full DB by one point, likely because removing redundant head-class samples reduces the imbalance. On UHCS with $IR = 10$, the most skewed federation, the baselines fall 9 to 23 points below Full DB at $p_f = 0.9$, while SCOPE stays within 1.4 points (Fig.~\ref{fig:params}b).
\par{Clients with the lowest accuracy:}
Large pruning also decides which clients lose accuracy (Fig.~\ref{fig:perf}b). A client whose data consists mostly of classes that are rare in the federation suffers when these classes are pruned. SCOPE does not prune classes that the global profile marks as rare. On UHCS, its 10\% worst clients reach 88.4\%, about two points above the best baseline, and on CIFAR-100 the margin is smaller.
\par{Rounds and computation to the target:}
On Tiny-ImageNet, SCOPE reaches 95\% of the Full DB accuracy about as fast as Full DB and 98\% in 25.5 fewer rounds (Table~\ref{tab:cost}), likely because the coreset is less imbalanced. Each round is also cheaper, since the sampled clients train on smaller datasets, so SCOPE reaches 98\% with 16\% less client computation than Full DB and the least of all methods, whereas the baselines spend more on their warm-up than their coresets save. On UHCS, SCOPE reaches 98\% of the Full DB accuracy within about ten rounds, as fast as Full DB and three times faster than FedCore, while FedCS and EL2N never reach it (Fig.~\ref{fig:perf}c). FedCore with the initial model, which Table~\ref{tab:cost} does not include, needs fewer rounds on Tiny-ImageNet and sends nothing before training, but its coreset falls behind SCOPE's under large pruning (Fig.~\ref{fig:perf}a). If the sampled clients run in parallel, a training round on Tiny-ImageNet would last about 2.5~s: 1.3 to 1.6~s for the client with the most data, about 1~s for aggregation, and 77~ms of transfer at Delta's line rate. The end-to-end time is therefore set mainly by the number of rounds, which SCOPE reduces by 23\% to reach 98\%, while its selection adds about 3~s per client.
\label{subsec:beta}
\par{Effect of $\beta$ and $\gamma$:}
$\beta$ decides which classes undergo redundancy pruning. On Tiny-ImageNet, $\beta = 0.9$ also prunes tail classes: with this margin, the 20\% rarest classes keep as little as 56\% of their samples (Fig.~\ref{fig:params}a). The rarity weight $\gamma$ limits this damage. At $\beta = 0.9$, the default $\gamma = 0.7$ keeps 11 to 18 more points of tail data than no weight, whereas at the default $\beta = 0.5$ the margin alone already spares most tail classes. On CIFAR-10, $p_l$ and $\gamma$ change accuracy by less than one point, and all experiments use the same defaults.
\subsection{Robustness, Generality, and Ablation}
\label{subsec:backbone}
\begin{table}[htbp]
\centering
\caption{Robustness of the training result on Tiny-ImageNet at $p_f = 0.5$ under different local training. Accuracy and global rounds to 98\% of the Full DB accuracy are given for Full DB / FedCS / SCOPE, where -- means not reached; both come from a separate set of runs that log every round, so accuracies differ slightly from Table~\ref{tab:acc_main}. Comp.\ is SCOPE's client computation over 200 rounds, selection included, relative to Full DB. The runs with $E = 20$ and FedProx use one seed.}
\label{tab:robust}
\scriptsize
\setlength{\tabcolsep}{2.5pt}
\begin{tabular}{l c c c}
\toprule
Configuration & Acc. (\%) & Rounds to 98\% & Comp. \\
\midrule
FedAvg, $E$=5     & 54.4 / 54.5 / 54.0 & 109 / 106 / 83.5 & $-8\%$ \\
FedAvg, $E$=20    & 47.9 / 45.7 / 47.6 & 95 / -- / 78     & $-21\%$ \\
FedProx, $E$=5    & 54.4 / 54.8 / 54.3 & 95 / 89 / 55     & $-14\%$ \\
\bottomrule
\end{tabular}
\end{table}
\par{Local epochs and FedProx:}
The benefit of the coreset holds when the training changes (Table~\ref{tab:robust}). With 20 local epochs, FedCS loses accuracy and never reaches 98\% of the Full DB accuracy, while SCOPE still needs fewer rounds than Full DB, 78 against 95. With FedProx, SCOPE needs 55 rounds against 95. In every configuration, SCOPE also needs 8 to 21\% less client computation than Full DB over 200 rounds, because each round trains on less data.
\par{GPU types, CPUs, and multiple GPUs:}
SCOPE's selection is about $1.8\times$ faster than FedCS's on all three GPU types we measured, because it runs one forward pass of a small encoder instead of training. It also runs without a GPU: a 32-core CPU node takes only $4.1\times$ longer than a GPU, so nodes attached to instruments can select their coresets on their CPUs. Because samples are scored independently, the selection also scales across GPUs: spreading the 100 clients over four GPUs of one node finishes it $4.0\times$ sooner, and over eight GPUs of two nodes $6.3\times$ sooner, with the same coreset (Table~\ref{tab:dist}). For the same reason, a client whose node has several GPUs can split its data into shards, score them in parallel, and merge their statistics exactly (Eq.~\eqref{eq:global_var}) before its single upload.
\begin{table}[htbp]
\centering
\caption{Wall-clock time of SCOPE's selection for all 100 TIN clients, 49{,}754 images, spread over 1, 4, or 8 GPUs; each GPU scores its clients one after another. The time covers loading the scorer, reading the images from the shared file system, scoring, exchanging the statistics, and pruning. Every configuration uploads 272~KB and keeps the same coreset.}
\label{tab:dist}
\scriptsize
\setlength{\tabcolsep}{4pt}
\begin{tabular}{lcccc}
\toprule
GPUs (nodes) & Clients per GPU & Wall (s) & Speedup & Efficiency \\
\midrule
1 (1) & 100 & 292.1 & $1.0\times$ & -- \\
4 (1) & 25 & 72.3 & $4.0\times$ & 101\% \\
8 (2) & 12--13 & 46.5 & $6.3\times$ & 79\% \\
\bottomrule
\end{tabular}
\end{table}
\begin{table}[htbp]
\centering
\caption{Data kept (\%) by SCOPE's selection on Tiny-ImageNet ($IR = 5$, $p_l = 0.1$): all classes / 20\% rarest classes, for the pruning rate $p_f$, the targeting margin $\beta$, and the rarity exponent $\gamma$ ($\gamma = 0$: rarity weight off). Bold: defaults.}
\label{tab:beta_gamma}
\scriptsize
\setlength{\tabcolsep}{2pt}
\begin{tabular}{cc cccc}
\toprule
$p_f$ & $\beta$ & $\gamma = 0$ & $\gamma = 0.35$ & $\gamma = 0.7$ & $\gamma = 1.4$ \\
\midrule
\multirow{3}{*}{0.5} & 0.3 & 80.6 / 88.5 & 80.9 / 89.0 & 81.2 / 89.0 & 81.9 / 89.0 \\
                     & 0.5 & 76.5 / 87.8 & 77.4 / 88.7 & \textbf{78.3 / 89.0} & 79.2 / 89.0 \\
                     & 0.9 & 57.0 / 69.6 & 58.7 / 75.7 & 60.6 / 80.9 & 63.4 / 87.7 \\
\midrule
\multirow{3}{*}{0.9} & 0.3 & 72.9 / 88.1 & 73.6 / 89.0 & 74.1 / 89.0 & 75.4 / 89.0 \\
                     & 0.5 & 65.6 / 86.9 & 67.3 / 88.5 & \textbf{68.8 / 89.0} & 70.4 / 89.0 \\
                     & 0.9 & 32.0 / 56.3 & 34.7 / 65.8 & 37.8 / 74.7 & 42.6 / 86.8 \\
\bottomrule
\end{tabular}
\end{table}
\begin{table}[htbp]
\centering
\caption{Accuracy (\%) of different architectures trained on one SCOPE coreset, selected once with MobileCLIP-S2, on Tiny-ImageNet with $IR = 5$ and $p_f = 0.7$. The baselines select their coresets with ResNet-50.}
\label{tab:arch_agnostic}
\scriptsize
\begin{tabular}{l ccccc}
\toprule
Method & ResNet-50 & Swin-Tiny & ViT-Tiny & DeiT-Tiny & EffNet-B0  \\
\midrule
GraND  & 53.48 & \textbf{67.19} & 43.71  & 41.48 &  35.20  \\
EL2N    & 53.35 & 66.76  & 44.22 & 43.79 &  35.48 \\
FedCS   & 54.90 & 66.85 &  45.05 & 43.35 &  36.71  \\
FedCore & 55.06 & 66.47 &  44.52 & 45.04 & 38.69  \\
\midrule
SCOPE  & \textbf{55.68} & 67.17 & \textbf{45.27}  & \textbf{46.48} & \textbf{38.72}   \\
\bottomrule
\end{tabular}
\end{table}\begin{table}[htbp]
\centering
\caption{Diagnostics of the two filters at $p_f = 0.5$ and $p_l = \eta$, one seed. Left, Consensus Filter with a fraction $\eta$ of labels flipped: precision and recall of its removals, and accuracy of FedCS / SCOPE. Right, Balancing Filter: median contraction $\rho_k$ of the client-to-global transport distance, and client drift before and after selection.}
\label{tab:noise}
\scriptsize
\setlength{\tabcolsep}{1.3pt}
\begin{tabular}{l c ccc c c cc}
\toprule
& & \multicolumn{4}{c}{Label noise} & & \multicolumn{2}{c}{Transport and drift} \\
\cmidrule(lr){3-6} \cmidrule(lr){8-9}
Setting & $\eta$ & Prec. & Rec. & Acc. & $\Delta$ & & Med.\ $\rho_k$ & Drift, raw $\to$ core \\
\midrule
\multirow{2}{*}{CIFAR-10, $IR{=}10$} & 0.1 & 59.4 & 58.5 & 37.73 / \textbf{42.26} & $+4.5$ & & \multirow{2}{*}{0.95} & \multirow{2}{*}{15.38 $\to$ 7.53} \\
                                     & 0.2 & 66.1 & 67.0 & 35.67 / \textbf{38.86} & $+3.2$ & & & \\
\multirow{2}{*}{UHCS, $IR{=}10$}     & 0.1 & 18.0 & 12.9 & 87.43 / \textbf{88.25} & $+0.8$ & & \multirow{2}{*}{1.09} & \multirow{2}{*}{1.36 $\to$ 1.50} \\
                                     & 0.2 & 32.7 & 26.5 & \textbf{73.77} / 73.50 & $-0.3$ & & & \\
\bottomrule
\end{tabular}
\end{table}
\par{One coreset for every architecture:}
Because SCOPE's scorer is frozen, one coreset serves every architecture. Selected once, it gives the best accuracy on four of five architectures and is within 0.02 points of the best on Swin-Tiny (Table~\ref{tab:arch_agnostic}). A model-based baseline must instead select again for every training model, since its scores come from that model. Providing coresets for three models therefore costs the baselines 3.7 to $24\times$ SCOPE's single selection, most of it for the larger models on $224 \times 224$ inputs (Fig.~\ref{fig:generality}a). When FedCS selects with each training model, it matches SCOPE's accuracy within 0.6 points, but only at this repeated cost.
\par{Size of the scorer backbone:}
A larger scorer does not give a better coreset (Fig.~\ref{fig:generality}b). MobileCLIP-S2 needs 79\% less GPU memory than ViT-H-14 and selects several times faster, yet it comes within 0.06 points of the best backbone on Tiny-ImageNet and is the best on UHCS.
\par{Sensitivity to prompts and class names:}
SCOPE does not depend on careful prompts. With a single prompt template, or with neutral tokens such as ``class 1'' instead of the class names, accuracy stays within 0.6 points of the default on CIFAR-10 and UHCS, although the pruned samples overlap little with the default ones. Which classes are pruned, and by how much, thus matters more than which samples within a class are removed; Tiny-ImageNet uses neutral tokens throughout.
\par{Removing the profile or a filter:}
\label{subsec:ablation}
The global profile matters most, and it costs only one exchange of statistics. Without it, clients discard samples of classes that are abundant locally but rare in the federation, which costs up to 24 points on CIFAR-10 at $p_f = 0.9$ (Fig.~\ref{fig:generality}c,d). Removing either filter costs about 3 points. Replacing the redundancy score by $2\hat{Z}_{RS,i} - \hat{Z}_{S_{neg},i}$ prunes nearly the same samples, which confirms that the diversity score acts as a reweighting of $RS$ (Section~\ref{subsec:metric_design}).
\subsection{Effect of the Filters on Label Noise and Client Drift}
\label{subsec:theory_validation}
Section~\ref{sec:convergence} gives two conditions under which FedAvg on the coreset approaches the clean objective: the Consensus Filter must remove the samples whose gradients deviate most, typically mislabeled ones (Assumption~\ref{ass:semantic}, Lemma~\ref{lemma:bias_reduction}), and the Balancing Filter must bring each client's data closer to the global data, $\rho < 1$ in Lemma~\ref{lemma:drift_reduction}, which reduces client drift. Table~\ref{tab:noise} tests both. On CIFAR-10 with flipped labels, the Consensus Filter works well: about 60\% of the samples it removes are mislabeled, three to six times the random rate, so SCOPE beats FedCS by up to 4.5 points. On UHCS the filter finds fewer flipped labels, since micrographs differ from the scorer's pretraining data, and the two methods are within one point. For the Balancing Filter, the median contraction is $\rho_k = 0.95$ on CIFAR-10, and client drift halves from 15.38 to 7.53. On UHCS, $\rho_k = 1.09$ and drift rises slightly from 1.36 to 1.50.
\subsection{Limitations}
\label{sec:limitations}
We assume a federation of identical GPU nodes, so our evaluation does not cover heterogeneous HPC clients, whose slower nodes would lengthen selection and each training round. SCOPE also relies on a pretrained vision-language model, and adapting it to new scientific domains, whose images differ from its pretraining data, remains a challenge.
\section{Conclusion}
SCOPE shows that a federation can select its coreset without training. Clients score their data once with a frozen vision-language model, so selection costs the same for any training model and grows less with the federation than model-based selection. A single exchange of per-class statistics gives every client a global view, replaces the warm-up rounds of model-based methods, and lets clients prune common classes while keeping rare ones. The resulting coreset is the most accurate in most settings, especially under large pruning and strong imbalance, and reaches 98\% of the full-data accuracy in 23\% fewer rounds than full-data training. Training-free selection is therefore a practical way to lower the cost of federated learning across scientific HPC nodes.
\FloatBarrier
\section*{Acknowledgments}
This research is supported by the U.S.\@ Department of Energy (DOE) through the Office of Advanced Scientific Computing Research's ``Orchestration for Distributed \& Data-Intensive Scientific Exploration'' and the ``Decentralized data mesh for autonomous materials synthesis'' AT SCALE LDRD at Pacific Northwest National Laboratory, which is operated by Battelle for the DOE under Contract DE-AC05-76RL01830.  We also appreciate the support from the Iowa State University Dean’s Emerging Faculty Leaders Award. We also thank the National Center for Supercomputing Applications for providing GPUs through allocation CHE240191 from the Advanced Cyberinfrastructure Coordination Ecosystem: Services \& Support (ACCESS) program.

\end{document}